\documentclass[conference]{IEEEtran}
\IEEEoverridecommandlockouts
\usepackage{cite}
\usepackage{amsmath,amssymb,amsfonts}
\usepackage{algorithmic}
\usepackage{graphicx}
\usepackage{textcomp}
\usepackage{xcolor}
\usepackage{placeins}
\usepackage{float}
\def\BibTeX{{\rm B\kern-.05em{\sc i\kern-.025em b}\kern-.08em
    T\kern-.1667em\lower.7ex\hbox{E}\kern-.125emX}}
\begin{document}

\title{Combating high variance in Data-Scarce Implicit Hate Speech Classification}

\author{\IEEEauthorblockN{Debaditya Pal}
\IEEEauthorblockA{\textit{Computer Science and Engineering} \\
\textit{ABV-IIITM}\\
Gwalior, India \\
debaditya.pal6@gmail.com}
\and
\IEEEauthorblockN{Kaustubh Chaudhari}
\IEEEauthorblockA{\textit{Computer Science and Engineering} \\
\textit{ABV-IIITM}\\
Gwalior, India \\
ckaustubhm06@gmail.com}
\and
\IEEEauthorblockN{Harsh Sharma}
\IEEEauthorblockA{\textit{Computer Science and Engineering} \\
\textit{ABV-IIITM}\\
Gwalior, India \\
harshhsharma23@gmail.com }
}

\maketitle

\begin{abstract}
Hate speech classification has been a long-standing problem in natural language processing. However, even though there are numerous hate speech detection methods, they usually overlook a lot of hateful statements due to them being implicit in nature. Developing datasets to aid in the task of implicit hate speech classification comes with its own challenges; difficulties are nuances in language, varying definitions of what constitutes hate speech, and the labor-intensive process of annotating such data. This had led to a scarcity of data available to train and test such systems, which gives rise to high variance problems when parameter-heavy transformer-based models are used to address the problem. In this paper, we explore various optimization and regularization techniques and develop a novel RoBERTa-based model that achieves state-of-the-art performance.
\end{abstract}


\section{Introduction}

In the current era of growing online communication, hate speech is prevalent. Sometimes, this hate speech may be implicit in nature, i.e. it may not contain expletives or explicit language. In such cases, the hate is delivered through sarcasm \cite{waseem-hovy-2016-hateful}, humor \cite{10.1145/3232676}, or some other literary device \cite{sanguinetti-etal-2018-italian}. This makes it a difficult task to detect such hate speech since most models are trained on datasets that focus specifically on expletives \cite{basile-etal-2019-semeval}, explicit phrases \cite{Silva_Mondal_Correa_Benevenuto_Weber_2021} or racial epithets \cite{warner-hirschberg-2012-detecting}. Implicit hate speech detection is hence a more challenging and nuanced problem, and there has been a lot of progress in recent years in this domain \cite{sap-etal-2019-risk} \cite{kennedy_et_al} . The reason implicit hate speech detection is challenging is that there is no clear definition for what does and does not constitute hate speech \cite{https://doi.org/10.48550/arxiv.2205.01374}. Furthermore, detecting hateful intent is difficult because of nuances in the language, the context of the conversation, and the tone of speech. Finally, data sources for implicit hate speech detection are scarce, and so modern natural language processing methods such as transformers (BERT, RoBERTa, GPT etc) tend to heavily overfit the dataset. This results in a severe high variance problem, which makes the implicit hate speech detection task even more challenging.

Our approach has been mainly aimed at reducing the effects of high variance during training. 
\begin{itemize}
    \item We experimented with layer-wise learning rate decay \cite{revisit-bert-finetuning} in order to preserve the low-level features of the pretrained model and better fine-tune the high-level features.
    \item We replaced dropout with mixout regularization in our model as it has proven to be a more efficient regularization method for pretrained language models \cite{https://doi.org/10.48550/arxiv.1909.11299}.
    \item We experimented with the weights of the last few layers of the language model, including weight reinitialization \cite{dodge2020fine} as well as concatenation and averaging the weights \cite{https://doi.org/10.48550/arxiv.1905.06316} before passing the embeddings onto the classification head.
\end{itemize}

We have presented a comparative study of the application of these methods to reduce variance and have developed a novel model that achieved state-of-the-art performance in the implicit hate detection task introduced in \cite{elsherief-etal-2021-latent}.

\section{Related Work}
Hate speech detection has been an active area of research in natural language processing for a long time. However, it is only recently that the field of implicit hate detection has been given the attention it deserves. Data scarcity has been a significant problem in this field but \cite{elsherief-etal-2021-latent} has curated a high quality human annotated dataset from public Twitter data. This dataset contains three stages. The first stage deals with a coarse grained binary classification problem. The second stage provides more details for the Tweets tagged as hateful in the first stage. This stage introduces six-way fine grained labels providing valuable information on the type of hate being peddled through the post. The third stage of the dataset also deals with implicitly hateful Tweets and provides information on the group being targeted by the hate speech and what the actual implied statement is. The paper also provides a benchmark for the task utilizing both support vector machines as well as transformer-based encoders such as BERT. We are concerned with the first two stages only as the third stage is not a classification problem.

\section{Implicit Hate Classification}
In this section, we review the structure of the problem we are trying to solve, the discriminative fine tuning strategies that helped us design a novel transformer based model as well as the architecture of the model itself.

\subsection{Task and Data Description}

\subsubsection{Stage 1: Binary Classification}
The first stage of the implicit hate corpus introduced in \cite{elsherief-etal-2021-latent} contains 19,112 natural language utterances, out of which 933 are labelled as "explicit hate", 4,909 are labelled as "implicit hate" and 13,291 are labelled as "not hate". However, the task assigned to this stage by the authors is that of binary classification between the "implicit hate" and "not hate" classes. The dataset is skewed towards the "not hate" class which is an accurate representation of the data distribution in real life. Hence, in order to address the skewness in the data we have used a weighted Cross Entropy Loss function.

\begin{equation} \label{eq:1}
    l_n = - \sum_{c=1}^{C} w_c log\frac{exp(x_{n,c})}{\sum_{i=1}^{C} exp(x_{n,i})} y_{n,c} 
\end{equation}
 where \(x\) is the input, \(y\) is the target, \(w\) is the weight, \(C\) is the number of classes.

\subsubsection{Stage 2: Fine Grained Classification}
This stage further annotates the 4,909 natural language utterances that were previously tagged as "implicit hate" in Stage 1. The utterances are further grouped into six categories, namely: Grievance, Incitement, Inferiority, Irony, Stereotypical and Threatening. Upon further human annotation it was found out that certain utterances which was previously tagged as "implicit hate" actually belong to the "not hate" category. After removal of these noisy samples an extreme class imbalance was introduced in this stage, which is why the authors expanded the dataset to 6,343 Tweets using bootstrapping and out-of-domain samples so that classifier performance is not severely hampered. That is why we have used an unweighted Cross Entropy Loss function for this stage as the corpus skewness is already dealt with. The equation is the same as Equation \ref{eq:1} but lacks the weight parameter.

\subsection{Model Architecture}
Transformer based encoder models have been shown to outperform linear support vector machines \cite{10.1145/3447535.3462484} and RNN based models \cite{mathew2020hatexplain} in hate detection tasks. Therefore we have decided to use RoBERTa \cite{https://doi.org/10.48550/arxiv.1907.11692} as our encoder. RoBERTa was pretrained on a much larger corpus compared to BERT and uses dynamic masking during language modelling. It has outperformed BERT in several NLP related tasks which serves as our motivation to choose it instead of BERT. 

RoBERTa expects tokenized input sequences marked with a few special tokens to signify the start and end of each sequence and it outputs context aware embeddings for these sequences.

These embeddings are then fed into a multi-layer perceptron network with two hidden layers each containing 100 neurons with the hyperbolic tangent activation function between them. We have tried to match our model architecture with that of \cite{elsherief-etal-2021-latent} as much as possible in order to remove architecture based advantages that may mislead the inference of the conducted experiments.

\subsection{Fine Tuning Strategies}
\subsubsection{Layer-wise Learning Rate Decay}
RoBERTa consists of 12 transformer block layers, each generating their own embeddings and capturing different semantic aspects of the input sequence. The earlier layers extract low level features and the later layers build up on that knowledge. Thus, it was proposed in \cite{howard-ruder-2018-universal} that different layers should be fine tuned till different extents. Using the same learning rate throughout the model risks a drastic change of the low level features that the pretrained language model was. This could lead to overfitting of a particular training set as the model learns to extract features relevant only to that training set and loses its generalization.

We have created three different setups for experimentation. In the first, we use the same learning rate and learning rate decay throughout the entire model, this serves as our baseline. In the second setup, we create a distinction between the RoBERTa encoder and the classification head and used different parameters for both of them. In the third setup, we create disctinctions within the RoBERTa encoder as well and create the four groups as stated in Table \ref{tab:GLLRD}. The grouping was done as follows: group 1 consisted of the embeddings, and layers 0-3 of the model, group 2 consisted of layers 4-7 and group 3 consisted of layers 8-11. 
\begin{table}
    \centering
    \begin{tabular}{|l|l|}
        \hline
        \textbf{Learning Rate = lr}  & \textbf{Group Learning Rate} \\ \hline
        \textbf{Group 1}             & lr / 2.6                     \\ \hline
        \textbf{Group 2}             & lr                           \\ \hline
        \textbf{Group 3}             & lr * 2.6                     \\ \hline
        \textbf{Classification Head} & lr * 10                      \\ \hline
\end{tabular}
    \smallskip
    \caption{Group Wise Learning Rates}
    \label{tab:GLLRD}
\end{table}

\begin{table*}
\centering
\resizebox{\textwidth}{!}{
\begin{tabular}{|l|l|l|l|l|}
\hline
\textbf{Model}                                                & \textbf{Precision} & \textbf{Recall} & \textbf{Accuracy} & \textbf{F-Score} \\ \hline
BERT \cite{elsherief-etal-2021-latent}                                                          & 72.1               & 66.0            & 78.3              & 68.9             \\ \hline
BERT + Aug \cite{elsherief-etal-2021-latent}                                                   & 67.8               & 73.2            & 77.5              & 70.4             \\ \hline
RoBERTa Baseline                                                      & 65.26              & 73.19           & 77.02             & 69.00            \\ \hline
RoBERTa + LLRD(2-Groups)                                      & 70.60              & 70.10           & \textbf{79.35}    & 70.35            \\ \hline
RoBERTa + LLRD(4-Groups)                                      & 70.32              & 70.52           & 79.30             & 70.42            \\ \hline
RoBERTa + LLRD(4-Groups) + Re-init(3)                         & 64.97              & \textbf{77.05}  & 77.46             & 70.49            \\ \hline
RoBERTa + LLRD(4-Groups) + Re-init(3) + Mixout(0.7)           & \textbf{71.17}     & 71.23           & 70.21             & \textbf{71.20}   \\ \hline
RoBERTa + LLRD(4-Groups) + Avg Last 4 Layers + Mixout(0.7)    & 68.11              & 72.56           & 78.54             & 70.26            \\ \hline
RoBERTa + LLRD(4-Groups) + Concat Last 4 Layers + Mixout(0.7) & 70.14              & 69.40           & 78.98             & 69.77            \\ \hline
\end{tabular}}
\smallskip
\caption{\label{Task 1}
Binary Classification of Implicit Hate Dataset
}
\end{table*}

\begin{table*}[t]
\centering
\resizebox{\textwidth}{!}{
\begin{tabular}{|l|l|l|l|l|}
\hline
\textbf{Model}                                                & \textbf{Precision} & \textbf{Recall} & \textbf{Accuracy} & \textbf{F-Score} \\ \hline
BERT \cite{elsherief-etal-2021-latent}                       & 59.1 & 57.9  & 62.9  & 58.0           \\ \hline
BERT + Aug \cite{elsherief-etal-2021-latent}                 & 58.6 & 59.1  & 63.8  & 58.6          \\ \hline
RoBERTa Baseline                                              & 64.49 & 64.43 & 64.51 & 64.35            \\ \hline
RoBERTa + LLRD(2-Groups)                                      & 63.74 & 64.11 & 63.95 & 63.90            \\ \hline
RoBERTa + LLRD(4-Groups)                                      & 64.79 & 64.62 & 64.43 & 64.68            \\ \hline
RoBERTa + LLRD(4-Groups) + Re-init(2)                         & 64.60 & 64.72 & 64.35 & 64.72            \\ \hline
    RoBERTa + LLRD(4-Groups) + Re-init(2) + Mixout(0.7)           & \textbf{65.90} & \textbf{65.26} & \textbf{65.47} & \textbf{65.58}   \\ \hline
RoBERTa + LLRD(4-Groups) + Avg Last 4 Layers + Mixout(0.7)    & 63.46              & 63.79           & 63.31             & 63.16            \\ \hline
RoBERTa + LLRD(4-Groups) + Concat Last 4 Layers + Mixout(0.7) & 64.81              & 65.13           & 64.91             & 64.90            \\ \hline
\end{tabular}}
\smallskip
\caption{\label{Task 2}
Six-way Fine Grained Classification of Implicit Hate Dataset
}
\end{table*}

\subsubsection{Mixout Regularization}
RoBERTa mainly uses weight decay and dropout regularization. In \cite{https://doi.org/10.48550/arxiv.1909.11299}, the authors proposed a technique called mixout regularization that was proved to be a more efficient form of regularization for large scale language models. Mixout regularization mainly replaces dropout and is applied on the fully connected layers. The idea behind mixout stems from the observation that setting neuron activations to 0, as is done is dropout, leads to a drastic loss of knowledge. Hence, mixout randomly selects neurons in a layer and replaces their parameters with that of the pretrained model. This helps to preserve the pretrained knowledge of the model and avoids the optimization from diverging away from the pretrained model.
\subsubsection{Re-initialization of Last Layers}
The process of fine tuning a language model is a trade-off between scope of the model's domain and its performance. In our case we limit the model's domain to mainly that of implicit hate detection but we gain an improved performance in that domain. Pretrained language models are highly generalized and hence, some of the high level features extracted by these models are not relevant to the domain of interest. Thus, following the general practice of transfer learning and the experiment proposed in \cite{revisit-bert-finetuning}, we re-initialize a few of the topmost layers of RoBERTa.

In this case, the number of layers to re-initialize becomes a hyper parameter and the performance  of the model is sensitive to this value.

\subsubsection{Using Intermediate Layer Embeddings}
Each of the 12 layers in RoBERTa generates its own embeddings that captures different semantic features of the input sequence. However, usually only the final layer pooler output is used for further computation. We have created two experimental setups to utilize the embeddings generated by the intermediate layers. In the first, we simply average the last 4 layers' embeddings and feed it to the classifcation head, this serves as a skip connection for the intermediate layer embeddings. In the second setup, we concatenate the embeddings from the last 4 layers with the motivation of preserving the embeddings of each individual layer.

Additionally we isolate this experiment from the weights reinitialization experiment as the intermediate layer embeddings only provide useful information if they are pretrained to some extent.

\FloatBarrier
\section{Experimental Setup}
A 60-20-20 split was used for training, validating and testing the RoBERTa based model. A linear learning rate scheduler with 10\% of the total training steps as warmup was used throughout all experiments. Three setups were followed in the layer-wise learning rate decay experiment. The first setup consisted of a uniform learning rate and decay for all layers, the second setup made distinctions between the RoBERTa model and the classification head and the third setup consisted of groups within the RoBERTa model. This model was then trained with a learning rate of \{1e-5, 3e-5, 5e-5\}, for 3 epochs with a batch size of 8. In the re-initialization experiment we reinitialized the weights of the last \{0, 1, 2, 3\} layers. The weights were reinitialized as per values from the normal distribution. The bias was reinitialized to zero. Next, we initialized mixout regularization with a mixout probability of \{0.3, 0.5, 0.7\}. Furthermore, the embeddings of the last four layers were also concatenated and averaged together (in separate experiments) when the weights were not reinitialized. 

\section{Results}
The results of the stage 1 experiments have been recorded in Table \ref{Task 1} and those of stage 2 have been recorded in Table \ref{Task 2}. For stage 2, the metrics have been calculated using the macro averaging method.

In stage 1, the RoBERTa baseline narrowly outperforms the BERT baseline in terms of F-score. However, upon adding the discriminative fine tuning and regularization techniques the performance improves significantly allowing us to create a new state-of-the-art benchmark with 2.3\% absolute improvement.

In stage 2, using RoBERTa instead of BERT already gives us a substantial boost in performance but as we observed, this performance  can be further improved by using the proposed techniques. After proper fine tuning of the hyperparameters we have been able to develop a model that outperforms the previous benchmark by 6.98\% in terms of F-score.

Utilizing the intermediate layer embeddings did have a slight improvement when compared to the RoBERTa baseline, however, the re-initialization experiment model outperformed this one in both stages.
\section{Conclusion and Future Work}
In our work, we have addressed the problem of high variance while fine-tuning RoBERTa, which is a large scale language model, on the implicit hate dataset which is relatively a small corpus. We have explored Layer-wise Learning Rate Decay as a discriminative fine tuning strategy along with efficient regularization techniques like Mixout. To better ensure domain transfer, we also experimented with re-initialization of few of the last layers of the encoder. Additionally, we sought to utilize the intermediate layer embeddings of RoBERTa to obtain a semantically richer contextualized embedding. The models developed with these techniques have been proven to have superior performance as compared to the previous benchmarks.

In future, these techniques can be combined with data augmentation as both have been proved to result in improved performance of the models. The reason behind the intermediate layer embedding models not performing at par with the re-initialized models can also be further investigated into, along with comparisons with other regularization and optimization techniques. There is still a large scope of research in the field of stabilizing the fine-tuning of large scale language models in data-scarce environments.

\bibliography{custom}
\bibliographystyle{IEEEtran}

\end{document}